# Making a Case for Social Media Corpus for Detecting Depression

Adil E. Rajput[1] and Samara M. Ahmed[2]

[1]College of Engineering, Effat University, Jeddah.
[2] College of Medicine, King Abdulaziz University, Jeddah.

**Abstract.** The social media platform provides an opportunity to gain valuable insights into user behaviour. Users mimic their internal feelings and emotions in a disinhibited fashion using natural language. Techniques in Natural Language Processing have helped researchers decipher standard documents and cull together inferences from massive amount of data. A representative corpus is a prerequisite for NLP and one of the challenges we face today is the non-standard and noisy language that exists on the internet. Our work focuses on building a corpus from social media that is focused on detecting mental illness. We use depression as a case study and demonstrate the effectiveness of using such a corpus for helping practitioners detect such cases. Our results show a high correlation between our Social Media Corpus and the standard corpus for depression.

## 1 Introduction

The ubiquitous nature of social media provides a mechanism for users of all ages and background to express their opinions. Such opinions are reflective of the user mental thought process and provide a window to underlying mental issue the end user might be suffering from. Given the stigma associated with mental illness globally -the stigma varies across different cultures- many individuals do not feel comfortable expressing the issues they face on a daily basis. The social media platform however, provides the users an opportunity to express their opinions anonymously and allow them to be more open and uninhibited. Such uninhibited and honest opinions can prove to be very valuable to practitioners in both the medical and social sciences fields (especially psychiatrists and psychologists). Another valuable factor is the fact that social media platforms allow the users to interact with various ecosystems. For example, a user can interact with a small fully connected social group (say friends from the same dormitory) or with a diverse group comprised of strangers mostly (say expressing opinion on a trending hashtag on twitter). This provides a window as to how the behaviour might change across the diverse ecosystems.

Major Depressive Disorder (MDD) simply known as depression is characterized by more than two weeks of persistent feeling of sadness and loss of interest [31]. Globally, 300 million people suffer from this disorder and certain cases can lead to suicide [32]. The American Psychiatric Association (APA) uses the DSM guidelines [30] that describes the criteria to detect depression. The professionals use certain criteria based on the patient's feedback to detect depression symptoms. Detecting depression among children and teenagers also have specific guidelines. Two major obstacles stand in the way of detecting depression namely 1) Lack of understanding the underlying symptoms on part of the patient and 2) the associated stigma.

The advances in the Big Data realm have provided new avenues to practitioners in various fields such as marketing etc. While certain fields have seen significant progress (e.g., one-on-marketing), the fields of psychology and psychiatry have yet to realize the full potential. Characterized by the 3Vs (Velocity, Veracity and Velocity) as described in [18], big data management allows the practitioners to sift through multitude of data and group related facts together to aid in the inference process. The Big Data discipline is an amalgamation of many research areas such as machine learning, information retrieval, distributed processing and natural language processing. When working with text data, the concepts of information retrieval play a key role. Specifically, they help in gathering various social media texts across the globe and allow machine learning algorithms to either cluster the related data together -unsupervised learning- or bring data together that fits a certain criteria -supervised learning (described in section 3).

The choice of corpora plays a significant role in scavenging the text data in various forms across different platform and retaining the data that pertains to the problem at hand. A corpus is defined as "A collection of naturally occurring text, chosen to characterize a state or variety of a language" [26]. The choice of corpora is important as it serves as the reference point in recognizing different words and eventually linking them together. As an example, [35] emphasized the need for having a corpus of American English in addition to one for British English as it will help in recognizing documents containing the prevalent American English vocabulary. The social media platform further complicates the issue with Out of Vocabulary (OOV) words that keep cropping up e.g., "whazzup w dat".

### 1.1. Problem Statement

In this paper, we address the following:
1. We build a standard corpus from standard essays and documents
2. We build a sample corpus from a set of hashtags that talk about depression (training data). We identify the keywords from the hashtags that will serve as the corpus and form the basis for searching across the social media platform
3. We validate the social media corpus against the standard corpus

## 2 Literature Review

Perhaps the oldest work in this area can be traced back to the seminal paper by Luhn [1] where he showed that the term frequency is an indicator of the concept underlying a document. This was bolstered by the inverse document frequency concept described in [2] where the author established the need to filter out stop words such as "the", "and", etc. This led us to the areas of Natural Language Processing (NLP) and Information Retrieval. NLP. ELIZA [3] was the first system that mimicked the understanding of natural language and was based on pattern matching. Palmer [4] describes various techniques for pre-processing the text. Stanford also made the NLTK toolkit available, which aids in text normalization and dealing with corpus written in Python [5]. Porter's pioneer work in developing a stemmer rounds up the traditional approaches to NLP [6]. Most of the work in matching text queries to documents depended upon the "edit distance" as the queries never returned an exact match as described by [7] among others.

The maturity in the field of Information Retrieval and NLP led to many practical applications such as sentiment analysis of the writer and the subject based on naïve Bayesian classifier suggested by [8] and refined by many other researchers. This led to the work done in various social media platforms. Examples include detecting sarcasm [11], event notification [12], emergency situation detection [9] and Geolocation prediction among others [10].

The need of corpus has been established in the seminal work by Schvaneveldt et. al. where the authors establish the need of a context to resolve ambiguity in text [26].  The work was also applied to Information Retrieval with similar results [36]. Our work builds a corpus based on similar premise where we focus on extracting standard and Out Of Vocabulary (OOV) keywords. The concept is also similar to [27] where the authors establish the need to build a corpus specific to American English speaking communities to understand their behaviour better. We also keep the work described in [28] in mind which focuses on volume of corpus. The challenges we face are regarding the language are similar to the work described in [34] – also related to social media platforms [34].

The aforementioned work on social media dealt with Big Data techniques. Researchers started applying such techniques in the field of medicine. Fan et. Al. showed the opportunity Big Data provides in studying rare events [13]. Murdoch [21] established the need for applying the big data techniques to medical field among others [18]. Chen et. al. described a framework on applying big data in the field of psychology [22]. The field of psychiatry saw many initiatives recently spurred by the interest in Big Data. These include developing suicide risk algorithms, risk of dementia, substance abuse disorders, prescribing psychotropic drugs and studying cognitive impairment. Monreith et. al. summarizes the above and describes the ramifications Big Data is having in the field of psychiatry in [23]. Dechoudhry et. al. has laid foundation for work in applying such techniques to social media platform [26]. Specifically, they have demonstrated predicting depression in Twitter users given a set of users who have indicated prevalence of depression in their lives. Our work however, does not target or specify a set of known users but rather targets known hashtags and builds a corpus for use as a standard for detection of depression symptoms. Rajput et. al. provide an overview of work done in this area [19].

## 3 Key Concepts

As mentioned earlier, applying Big Data concepts to social science require a confluence of various concepts in the field of Computer Science. Before delving into the crux of this paper, this section lists the concepts that come into play. The concepts described below are present plentiful in Computer Science literature.

*Machine Learning (ML)* is a sub field of Artificial Intelligence – the ability of machines to acquire intelligence over a period of time that comes naturally to humans. Machine Learning uses statistical techniques to learn about the data at hand and build predictive models. The research in this area dates back to 1950s and application ranges from biological networks to social sciences recently. Machine Learning is focused towards supervised or unsupervised learning.

*Data Mining* is very similar in nature to ML but with a key difference namely that data mining processes massive amount of data and detects hidden pattern(s). The data mining also uses the concept of supervised and unsupervised learning and has become increasingly popular due to the Big Data revolution.

*Supervised Learning* requires a training set of data with a clear set of input mapped to a clear set of output. The goal in this case is to figure out a rule and/or an algorithm that ties the inputs to the outputs. Once this part is done, the algorithm is tested against a set of test data to test the effectiveness of the discovered algorithm. The work in this paper is an example of this technique.

*Unsupervised learning* does not have access to the training data set and the goal is to find similar hidden data and group it together. The goal could be as s simple as separating the data into two broad groups – known as classification problem or form various clusters of data – known as clustering problem. There are many other problems in this realm. One application to Psychiatry could be to find tweets by a particular age group across the globe and pool it together.

*Natural Language Processing (NLP)* is a major area in the field of Machine learning where the goal is to process natural human language and infer meaningful insights. While the field dates back to early 1950s as we saw in the previous section, researchers became increasingly interested with the advent of the World Wide Web. The Social media revolution has furthered this cause with certain challenges. These include the users not adhering to standard language constructs and vocabulary. Building a corpus is one of the key requirements in Natural Language Processing. Furthermore the more specialized a corpus, better are the chances of answering queries accurately [31]. Our work is a step in this direction to move towards building a representative corpus for Social Media.

*Information Retrieval (IR)* is the process of matching a particular query against a set of sources. As opposed to a database where data is matched exactly, information retrieval deals with approximate matching. The search engines on the internet are examples of information retrieval application where a user's query is run against a collection of documents and the closest matching documents are presented to the user. The two criteria used in IR are precision and recall specific to the relevance of the results. Precision deals with the fraction of true positives while recall measures the proportion of true positives returned. While IR can be applied to non-text documents such as images and videos, the main application of IR has been in the natural text area. One of the most used techniques in determining the subject of a particular document is term frequency (tf) and Inverse document frequency (idf). The tf measures the frequency of the terms present in the document and infers the subject/keywords describing the document. The idf factor focuses on eliminating the common used words such as prepositions, articles etc. allowing the tf factor to accurately represent the subject of the document at hand. While many other techniques have been proposed and tested, tf-idf algorithm is usually the starting point when dealing with corpora.

## 4 Experiment Setup

### 4.1 Assumptions

To evaluate the effectiveness of our corpus for social media, we focus on Twitter. Specifically, we narrow down the problem to identify the keywords specific to depression. We assume the following:
1. We do not differentiate between various types of depression (e.g., post-partum depression, teenage depression etc.)
2. We focus on corpora in the English language only and hence our results will only apply to English speaking community (However, we do not make a difference between native and non-native speakers)
3. We build our corpora using a given set of hashtags that specifically talk about depression. (We later confirm this assumption by checking against the keywords gleaned from various articles on depression)
4. The corpus built does not detect cases of pure depression from those that have a case of comorbidity (e.g., anxiety etc.)

### 4.2 Data Sources and Data Gathering

One of the biggest challenges when gathering data is ensuring the legality of using the data [22]. The data we gather comes from the following sources
1. Standard Corpus – Comprised of essays on Depression and provide a basis to confirm our third assumption
2. Set of known hashtags that talk about depression
3. Tweets written by particular users who have participated in the hashtags

We do not store any user credentials and anonymize them by giving fictitious identifiers.

### 4.3 Preprocessing and Processing Data

After gathering the tweets, we perform the following steps
1. Removing the stop words: This means removing words such as "for", "to" etc.
2. Vectorize the remaining words: In simple terms, we use the famous term frequency algorithm to extract the keywords. We will use the tf-idf algorithm implemented in python. For each word, we calculate the frequency of all the words
3. Once the data is gathered, we anonymize the userids for privacy purposes
4. We store the data in a NoSql database - MongoDB. This provides us the groundwork to do future work in this field

**4.4 Evaluation**

After setting up the experiment, we ran the following experiments
1. We wanted to confirm our assumption about the keywords in our selected hashtags. Specifically, the hashtags generated two sets of keywords namely 1) standard dictionary words that the users typed and 2) Out of Vocabulary words (OOV) that cannot be found in a dictionary. We compared the frequency of similar dictionary words in the hashtags against those in the standard corpora gleaned from Depression literature. The underlying assumption was that if the frequency of standards word in hashtag bears a high correlation then by associativity the OOV words might represent the prevalent usage in social media
2. Once we established the effectiveness of our base corpus we next used the corpus against the random ten users we chose who participated in our base set of hashtags. Since we kept track of the words that each particular user contributed toward our corpus, we removed those words from the corpus for this particular user
3. We create a vector of key words for the particular user posts against our corpus and see the commonality of the frequent terms

**5 Results and Discussion**

### 4.1 Keyword Identification for Standard Corpus

The very first thing we intended to do in this work was to identify the key words that were used in both the standard corpus and the social media. Utilizing the tf-idf algorithm, described earlier, we identified essays on Major Depressive Disorder, Depression and Psychotic Depression (The third choice was arbitrary in terms of various types of depression). We identified the top fifteen words shown in Table 1.

We found the following observations interesting:
1. The standard corpus we chose is geared towards general audience and hence the keywords identified were in line with the general audience comprehension
2. Certain words identified were not typical when it comes to the discussion regarding depression. The word "Long" in Table 1 is an example of this. However, when taken into context of depression, it becomes relevant as it is used to describe other relevant symptoms e.g., **long** bouts of insomnia

| **Keyword** |
|---|
| Pleasure |
| Sleep |
| Therapy |
| Activities |
| Treatment |
| Long |
| Abuse |
| Health |
| Identity |
| Life |
| Mood |
| Mental |
| Depression |
| Physical |
| Pressure |
| Childhood |

*Table 1 : Keywords identified from standard corpus*

### 4.2 Standard Corpus versus Social Media

Once we established a base vector of keywords, we chose the hashtags as our base set for Twitter namely #depression, #depressed and #feelingdown. The #feelingdown hashtag might not necessarily be indicative of long-term depression but showed many tweets that could indicate potential prevalence of long-term depression.

Next, we set the sample size of tweets gleaned from the hashtags to 100, 200, 500 and 1000. The following was worth noting:
1. The keywords identified from standard corpus showed a very high correlation to the words identified from the hashtags
2. For sample sizes of 100 and 200, the correlation was almost 100% for all the words
3. Once the size increased to 500 and 1000, certain words' frequency started going down e.g., sleep. However, the interesting trend was that for n = 100, the word insomnia started going up in frequency (not shown below) and was ranked very high for n = 1000. One possible explanation could be that as number of tweets go up, the users' choice of words become more narrow and accurate. e.g., "having difficulty with sleep" versus "insomnia". Both terms however could possibly point to the symptom of sleep disorder that can accompany depression.
4. Despite the non-standard language used on the Internet and Twitter, the keywords identified in both corpora were similar. This result was consistent with the findings of [2,34] where the authors emphasized the following: 1) the narrower the context, better the selection of keywords and 2) despite the presence of OOV words, the underlying text still conforms to the subject at hand
5. The corpus gleaned from hashtags can serve as a basis to classify tweets from a test data to help detect depression

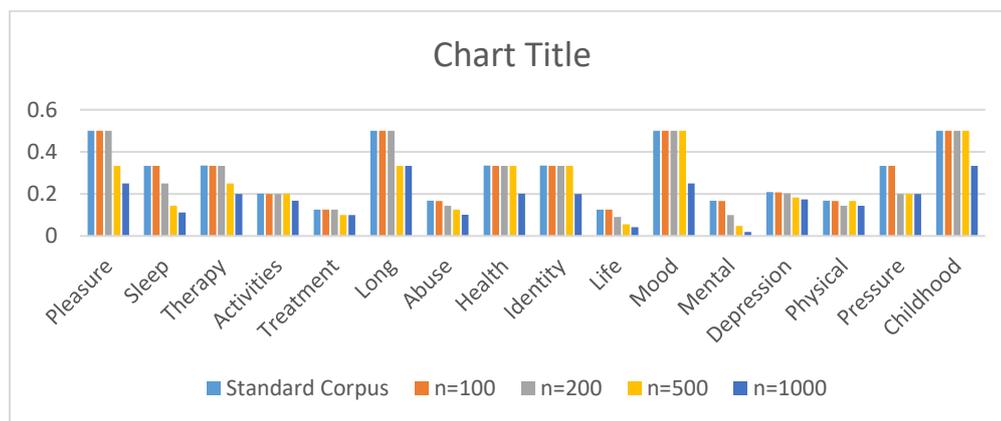

*Figure 1: Standard Corpus versus Hashtags Keywords*

## 6 Conclusion and Future Work

In this paper, we have demonstrated the need of establishing a corpus specific to social media in the context of mental health. Such a corpus will help the practitioners to detect detection for a particular disease. We used depression as our case study and demonstrated the high correlation between social media corpus and standard corpus. We also ran experiments by selecting certain users in Twitter, mined their tweets for a period, and showed a high correlation between the keywords they used and the corpus we built.

The work is the first in a series of steps in this area. Specifically, we would like to focus on the following:
1. Increase the size of the social media corpus and further validate it against the DSM V standards
2. Contribute towards refining the guidelines for use of Big Data in Psychiatry and Psychology
3. Test the corpus against different social media platforms and compare the results we obtained from Twitter
4. Compare the results obtained in English to another language and model the user behaviour